# Enhancing Type 2 Diabetes Diagnosis with Evolutionary Algorithms and Machine Learning

Parisa Rezaei
*Dept. of Computer Engineering and Information Technology*
*Graduate University of Advanced Technology*
*Kerman, Iran*
p.rezaei@student.kgut.ac.ir

Mohsen Saffar
*Dept. of Industrial Engineering*
*K. N. Toosi University of Technology*
*Tehran, Iran*
mohsen.saffar@email.kntu.ac.ir

Hamid Reza Naji*
*Dept. of Computer Engineering and Information Technology*
*Graduate University of Advanced Technology*
*Kerman, Iran*
naji@kgut.ac.ir

Mohammad Mehdi Faghih
*Dept. of Computer Engineering and Information Technology*
*Graduate University of Advanced Technology*
*Kerman, Iran*
m.faghih@kgut.ac.ir

Rasoul Nouriazar
*Dept. of Computer Engineering and Information Technology*
*Graduate University of Advanced Technology*
*Kerman, Iran*
r.nouriazar@student.kgut.ac.ir

***Abstract*—Type 2 diabetes is a widespread chronic disease that requires early and accurate diagnosis to prevent serious complications. Traditional diagnostic methods often lack sufficient accuracy, which highlights the need for more reliable computational solutions. In this study, we propose a model that integrates the Random Forest classifier with the Bat Optimization Algorithm for simultaneous feature selection and hyperparameter tuning. The SMOTE-ENN method was first applied to the Pima Indians Diabetes Database to correct class imbalance and remove noisy or ambiguous samples, producing a more balanced and cleaner dataset. The optimized model trained on this refined dataset achieved 89% accuracy, 88% precision, 90% recall, and an F1-score of 89%, clearly outperforming the baseline Random Forest and other existing approaches. These results demonstrate the potential of combining evolutionary algorithms with ensemble learning to provide a practical and cost-effective tool for early detection of type 2 diabetes in clinical practice.**

***Keywords—Diabetes, Diabetes Diagnosis, Machine Learning, Evolutionary Algorithms, Feature Selection***

## I. INTRODUCTION

Diabetes mellitus, an intricate and progressively common metabolic condition, is chiefly defined by persistent hyperglycemia arising from impairments in insulin production, insulin function, or a combination of both [1]. The extraordinary rise in the global prevalence of diabetes necessitates a comprehensive understanding of its nature, history, types, and management strategies, as the International Diabetes Federation indicates that approximately 463 million adults (aged 20-79) worldwide were living with diabetes in 2019, with projections suggesting this number will exceed 700 million by 2045 [2]. The historical roots of diabetes trace back to ancient civilizations, where descriptions resembling diabetes mellitus were first recorded in Egyptian medical texts, detailing the notable symptoms of excessive thirst and urination, which prompted the term "diabetes," derived from the Greek word for "siphon" [3].

Later developments in medical history further clarified the naming of diabetes. The term "mellitus," meaning "honey-sweet," was later appended in the second century AD by the Greek physician Aretaeus of Cappadocia when he observed that the urine of affected individuals was sweet due to high glucose levels, serving as a clear indication of the metabolic dysfunction inherent in this condition [3]. Recognized by physicians and scientists for centuries, the understanding of diabetes has evolved dramatically from myths and rudimentary explanations to the sophisticated biomedical insights of today, which delineate the mechanisms behind its pathogenesis, associated complications, and potential treatment options. diabetes significantly compromises patients' quality of life and predisposes them to acute complications that can necessitate hospitalization and further complex medical care [4].

As medical understanding has advanced, attention has increasingly turned toward modern technologies and lifestyle strategies in diabetes care. Over the years, research has illuminated the role of advanced technologies in managing diabetes, particularly through artificial intelligence (AI) and machine learning, which are increasingly used to predict diabetes-related complications and personalize treatment [5]. The integration of technology into diabetes care provides new avenues for continuous monitoring and tailored interventions, reflecting a shift towards more proactive and preventive healthcare practices [5]. Furthermore, lifestyle interventions, including nutritional adjustments, physical activity promotion, and the advent of digital health tools such as mobile apps, play a critical role in managing diabetes and preventing its progression [6].

In this study, we employ the Bat Optimization Algorithm to achieve optimal feature selection from the dataset and to optimize the parameters of the Random Forest machine learning model. Using data from the Pima Indians Diabetes Database, our goal is to build an effective and accurate model for detecting type 2 diabetes.

In the subsequent section, related concepts are explored and the literature is reviewed.

## II. REVIEW

Diabetes mellitus is a complicated and growing metabolic condition characterized by chronic hyperglycemia due to deficiencies in insulin secretion, insulin action, or both [1]. The effects of this disease on the world have grown a lot. The Published as: P. Rezaei, M. Saffar, H. R. Naji, M. M. Faghih, and R. Nouriazar, "Enhancing Type 2 Diabetes Diagnosis with Evolutionary Algorithms and Machine Learning," 2025 32nd National and 10th International Iranian Conference on Biomedical Engineering (ICBME), pp. 733-738, 2025. doi: 10.1109/ICBME68496.2025.11392543

International Diabetes Federation says that in 2019, almost 463 million adults (ages 20 to 79) have diabetes. Estimates say that by 2045, this number will be over 700 million [2]. Diabetes has been around for a long time, and early Egyptian medical literature talked about symptoms including excessive thirst and urination. The word diabetes comes from the Greek word for siphon [3]. The term "mellitus," meaning "honey-sweet," was added in the 2nd century AD by the Greek physician Aretaeus of Cappadocia after observing that the urine of affected individuals was sweet due to elevated glucose levels, a hallmark of the underlying metabolic dysfunction in this condition [3].

Building upon this historical and epidemiological perspective, it is essential to consider the classification of diabetes in order to better understand its underlying mechanisms and management. Diving deeper into the classification of diabetes, it is primarily divided into two main types: Type 1 diabetes mellitus (T1DM) and Type 2 diabetes mellitus (T2DM) [1]. T1DM is an autoimmune illness in which the immune system assaults and destroys beta cells in the pancreas that make insulin, which causes a complete lack of insulin. This type of diabetes usually shows up in children or teenagers, which is why it is called juvenile diabetes [7]. In contrast, T2DM, the more prevalent type, is characterized by insulin resistance and a relative deficiency of insulin, often linked to lifestyle factors such as obesity, sedentary behavior, and genetic predisposition, which highlights the multifactorial nature of its pathogenesis [1]. Approximately 85-90% of all diabetes cases worldwide can be attributed to T2DM, showing a troubling correlation with rising global obesity rates [2]. Additional categorization encompasses gestational diabetes mellitus (GDM), which manifests during pregnancy and generally subsides following childbirth, though it may elevate the risk of subsequent T2DM development for both the mother and the child [1]. GDM's prevalence has escalated in tandem with rising obesity levels among pregnant women, necessitating attentive prenatal care and lifestyle management [8]. The distinct pathophysiological mechanisms underlying these different forms of diabetes contribute to their varying clinical presentations and management strategies, with profound implications for patient care.

Understanding the forms of diabetes also necessitates recognizing their potential long-term consequences. The complications associated with diabetes mellitus are numerous and severe, impacting various organ systems due to chronic hyperglycemia and the resultant microvascular and macrovascular complications [1]. These include diabetic neuropathy, retinopathy, nephropathy, cardiovascular disease, and foot ulcers, whereby continuous high blood sugar levels lead to long-term damage to blood vessels and nerves [5]. Diabetes not only impacts daily life and well-being but also heightens the risk of sudden complications that may lead to hospital stays and the need for advanced medical care [4]. The wide array of diabetes-related complications underscores the importance of effective glycemic control and health management strategies to mitigate these risks and enhance patient outcomes.

Given the seriousness of these complications, timely and accurate diagnosis becomes a cornerstone of effective management. The global rise in diabetes highlights the urgent need for early and accurate diagnosis to reduce the burden of undiagnosed cases, which often lead to costly complications and reduced quality of life [9]. Early detection enables prompt intervention and significantly lowers the risk of long-term issues such as cardiovascular and kidney diseases, especially when diagnosis occurs at a younger age [10]. However, many individuals remain undiagnosed due to the subtle nature of early symptoms and widespread unawareness of risk factors, which contributes to delayed treatment and worsened outcomes [11]. Conventional diagnostic tools—like fasting plasma glucose, OGTT, and HbA1c—each have limitations in sensitivity and specificity, with HbA1c in particular sometimes failing to detect early-stage diabetes in certain populations [12]. To address these challenges, machine learning and artificial intelligence are increasingly used to analyze large datasets and detect patterns indicative of early disease, offering the potential for improved diagnostic accuracy [13]. Moreover, future directions include integrating technologies such as continuous glucose monitoring and telehealth platforms to enable real-time tracking and individualized care, ultimately supporting better disease management and prevention [14]. Public education also remains a cornerstone of early diagnosis efforts, with awareness campaigns shown to improve recognition of symptoms and encourage regular health screenings, particularly in high-risk and underserved communities [15].

In the identification of type 2 diabetes, multiple diagnostic criteria are commonly utilized, such as fasting plasma glucose (FPG), oral glucose tolerance test (OGTT), and glycated hemoglobin (HbA1c) assessments, along with others [16]. FPG, a widely used test, identifies diabetes at levels $\geqslant 126$ mg/dL after an 8-hour fast, offering a simple and accessible screening option, though it may be influenced by temporary physiological factors such as stress or illness [17]. OGTT, which measures glucose levels two hours after a glucose load, is more sensitive in detecting insulin resistance but is less practical for routine use due to its time-intensive nature [18]. HbA1c provides a longer-term view of blood glucose trends and is diagnostic at $\geqslant 6.5\%$, yet it may yield inaccurate results in individuals with conditions like anemia or hemoglobinopathies [19]. Additionally, real-time technologies such as continuous glucose monitoring (CGM) are emerging as valuable diagnostic tools, capable of capturing glycemic fluctuations that traditional tests may miss, thereby supporting more personalized care strategies [20]. Combining multiple diagnostic approaches may improve overall accuracy and help clinicians better assess both short-term and long-term glycemic control [21].

To further strengthen these conventional approaches, machine learning has introduced new possibilities. The integration of machine learning has significantly advanced the diagnosis and management of type 2 diabetes, offering new levels of accuracy in early detection and risk assessment [22]. Machine learning models, particularly those trained on large datasets from electronic health records, are effective in predicting complications and identifying high-risk patients before symptoms manifest [23]. Gradient boosting methods have also shown promise in distinguishing diabetic from non-diabetic individuals, supporting timely interventions [24]. Looking forward, research is increasingly focused on improving model explainability and ensuring ethical use of data to minimize bias in AI-driven healthcare applications [25].

To evaluate these advanced methodologies in practical applications, researchers have relied on conventional benchmark datasets. The Pima Indians Diabetes Database is a well-known dataset that is often used in investigations of diabetes diagnosis using medical data analysis and machine learning. This research employs this dataset. With 9 columns and 768 rows, it has 500 cases of people who don't have diabetes and 268 cases of people who do. The outcome variable is binary, with 0 for a negative diabetes test and 1 for a positive test. This publicly available resource can be accessed from Kaggle (https://www.kaggle.com/uciml/pima-indians-diabetes-database), and its detailed features are outlined in Table 1 [26].

TABLE 1. Summary of the Pima Indian diabetes dataset

| Feature | Description | Data type | Range |
|---|---|---|---|
| Preg | Number of pregnancies | Numeric | [0,17] |
| Gluc | Plasma glucose after 2-hour OGTT | Numeric | [0,199] |
| BP | Diastolic blood pressure (mmHg) | Numeric | [0,122] |
| Skin | Skinfold thickness (mm) | Numeric | [0,99] |
| Insulin | Serum insulin after 2 hours (µU/ml) | Numeric | [0,846] |
| BMI | Weight/height² (kg/m²) | Numeric | [0,67. 1] |
| DPF | Genetic risk score for diabetes | Numeric | [0. 078, 2. 42] |
| Age | Age in years | Numeric | [21, 81] |
| Outcome | Diabetes status (0 = non-diabetic, 1 = diabetic) | Factor | [0,1] |

Considering the intricacy of such datasets, feature selection stands out as an essential process for boosting model efficacy. In machine learning and data mining, feature selection holds significant importance, especially in the context of high-dimensional datasets, as it elevates model precision, diminishes computational demands, and bolsters interpretability through the removal of extraneous and redundant features [27]. To address these challenges, nature-inspired metaheuristic algorithms have been widely explored. The Bat Algorithm, introduced by Yang in 2010, mimics the echolocation behavior of microbats and employs adaptive control of frequency, loudness, and pulse emission rate to maintain a balance between global exploration and local exploitation. This strategy enhances solution diversity while accelerating convergence, proving effective for solving complex and nonlinear engineering optimization tasks [28]. In this study, this algorithm is employed to achieve optimal feature selection and to optimize the parameters of the Random Forest machine learning model.

The Random Forest algorithm is one of the most widely applied tree-based learning methods, in which multiple subsets of the observed data are randomly selected and a decision tree is constructed and trained for each subset [29]. As an ensemble machine learning model, Random Forest can be employed for both classification and regression tasks. In this approach, a subset of the training data is chosen at random, and several decision trees are generated [30]. Each tree independently performs classification, and the class receiving the majority of votes across the trees is considered the predicted class [31]. In other words, Random Forest is a parallel collection of decision trees that aggregate their outputs through majority voting or averaging to determine the final result, thereby reducing the risk of overfitting and improving predictive accuracy. For this reason, Random Forest models are generally more accurate than a single decision tree model [32].

In response to the increasing focus on feature selection and optimization for enhancing predictive accuracy, several recent investigations have examined diverse machine learning techniques applied to the PIMA Indian Diabetes dataset to evaluate their diagnostic efficacy.

Verma and Khatoon (2024) [33] analyzed applications of data mining for diabetes prediction using the PIMA Indian Diabetes dataset, applying Logistic Regression, Support Vector Machine (SVM), K-Nearest Neighbor (KNN), and Random Forest (RF), and evaluating performance through accuracy, precision, recall, and F-measure after preprocessing steps such as handling missing values and correlation analysis; their results showed RF achieved the highest accuracy (80.08%) and precision (86.53%), while KNN performed weakest, underscoring RF's superiority for diabetes diagnosis. Sreejith et al. (2020) [34] proposed a Clinical Decision Support System (CDSS) framework to address class imbalance and feature redundancy by integrating an enhanced SMOTE (Orchard-SMOTE) for data rebalancing and Chaotic Multi-Verse Optimization (CMVO) for feature selection; tested on three UCI datasets (Indian Liver Patient, Thoracic Surgery, and PIMA Indian Diabetes), their approach with RF classifiers achieved accuracies of 82.46%, 86.88%, and 89.04%, respectively, and statistical validation confirmed its effectiveness. Mushtaq et al. (2022) [35] developed a machine-learning framework for early diabetes prediction on the PIMA dataset, employing preprocessing steps such as SMOTE, Tomek links, and outlier removal, and testing multiple classifiers; RF achieved the highest single-model accuracy (80.7%), and an ensemble of RF, Naïve Bayes, and Gradient Boosting further improved accuracy to 81.7%. Anam et al. (2025) [36] introduced a diagnostic method combining RF with the Bat Algorithm (BA) for hyperparameter optimization, demonstrating that RF-BA outperformed standard RF and RF-PSO in accuracy, avoided overfitting, and required less computation time when evaluated on Kaggle diabetes datasets. Finally, Chang et al. (2023) [26] proposed an e-diagnosis system within the Internet of Medical Things (IoMT), applying Naïve Bayes, RF, and J48 to the PIMA dataset with preprocessing through imputation and feature selection; results indicated RF achieved the best overall accuracy (79.57%), Naïve Bayes performed well with fewer features (79.13%), and J48 achieved the highest sensitivity (~89%).

In the subsequent section, the proposed model is discussed in detail.

## III. METHOD

In this study, the Bat Optimization Algorithm was employed both for optimal feature selection and for tuning the Random Forest machine learning model, which served as the primary classification method on the Pima Indian Diabetes dataset. The proposed methodology consists of three main stages:

1. Initial preparation and cleaning of the Pima Indian Diabetes dataset.
2. Integrating optimal feature selection with the optimization of the Random Forest model in order to enhance classification performance. This stage leverages the Bat Optimization Algorithm in conjunction with the Random Forest classifier to improve the outcomes of the earlier steps.
3. Applying the Random Forest algorithm to perform classification on the dataset.

Figure 1 presents our suggested process as a simple model diagram. It basically maps out the research steps we followed to build the model.

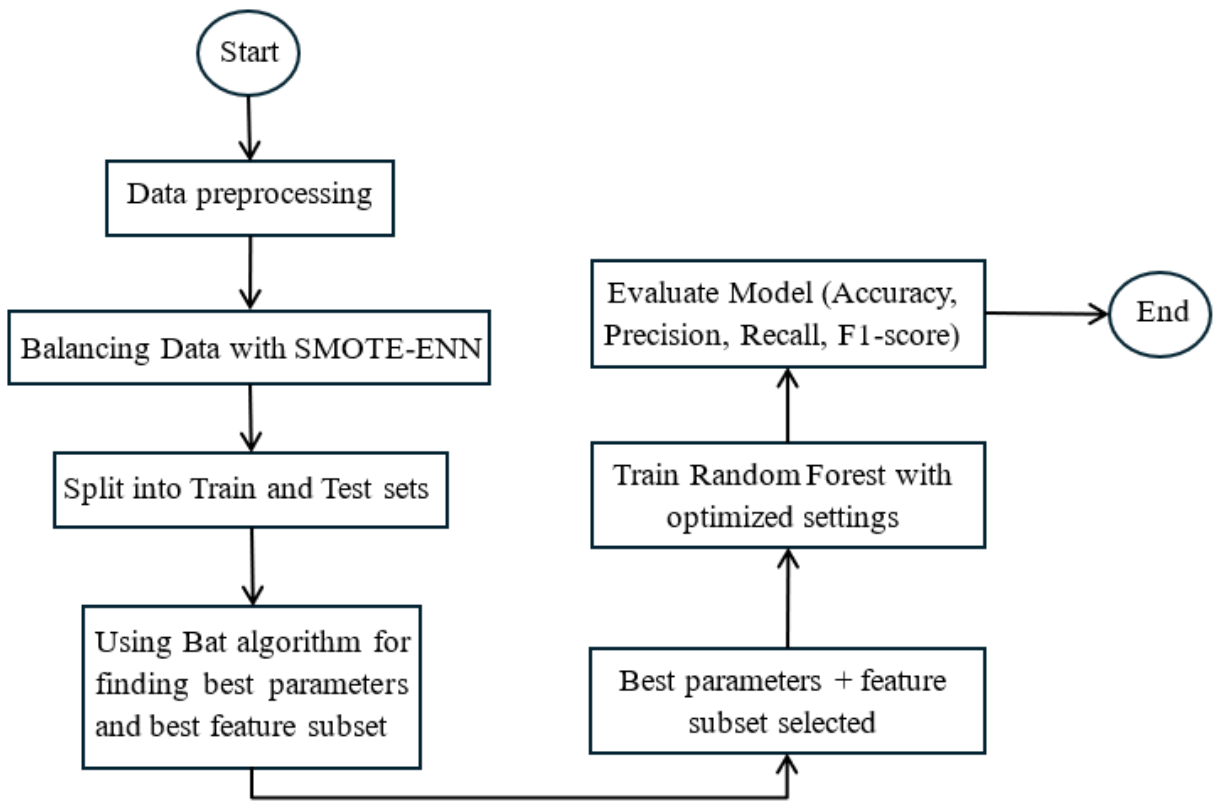

Fig. 1. Proposed Model Diagram

At the beginning of the process, the data must be preprocessed to enable their proper use in the model.

The Pima Indians Diabetes dataset does not contain explicit missing values (e.g., NaN or NULL), as all entries are populated. However, several clinical features—such as Glucose, BloodPressure, SkinThickness, Insulin, and BMI—include zero values that are physiologically implausible and are commonly interpreted as hidden missing data. In addition, the dataset exhibits outliers in variables such as Insulin, Glucose, BMI, Age, and BloodPressure. These anomalies often arise from human error, measurement inaccuracies, experimental conditions, sampling issues, or data-processing mistakes. Detecting and addressing such outliers is a critical step in the development of machine learning models, as they can substantially influence model performance [37]. For this reason, standardization or normalization of the data is recommended to mitigate the adverse effects of outliers and improve the robustness of predictive modeling.

To overcome these issues, the dataset was refined through two essential preprocessing steps. First, incomplete values within specific columns were carefully addressed and replaced, resulting in a complete and coherent dataset free of missing information and ready to support subsequent analyses. Following this, data points that exhibited patterns substantially different from the majority were identified as outliers, labeled accordingly, and removed. These steps ensured that the final dataset is both cleaner and more reliable, providing a solid foundation for accurate and meaningful analysis.

After completing these preprocessing steps, the model was then developed using a comprehensive, multi-stage approach designed to improve both data quality and predictive performance. At the outset, the challenge of class imbalance, which is common in real-world datasets and particularly in medical or diagnostic domains, was addressed using the SMOTE-ENN technique. This method combines synthetic oversampling with noise reduction, simultaneously generating new representative samples for minority classes while removing ambiguous or misclassified data points [38]. The outcome of this step is a more balanced and reliable dataset, which provides a stronger foundation for model training.

After obtaining a balanced and reliable dataset through preprocessing and class balancing, the process then advanced to feature selection and parameter optimization using the Bat Algorithm, a nature-inspired metaheuristic modeled after the echolocation behavior of bats. This algorithm was applied to search the high-dimensional space of possible feature subsets and model parameters, ultimately identifying both the most informative features and the optimal settings for the learning algorithm. Through this approach, unnecessary or less pertinent features were eliminated, whereas essential parameters such as the number of estimators and tree depth were optimized to improve model efficacy and prevent overfitting or underfitting.

Based on the optimized features and parameters identified in this stage, the final step involved training a Random Forest classifier on the refined dataset, ensuring that the model leveraged the most informative attributes under the best-performing configuration.

The confusion matrix for classifying diabetes-related data consists of four possible outcomes. True Positive (TP) occurs when a person with diabetes is correctly identified as diabetic. False Positive (FP) refers to the case where a healthy individual is incorrectly classified as diabetic. True Negative (TN) arises when a healthy individual is accurately recognized as non-diabetic. Finally, False Negative (FN) represents the situation in which a person with diabetes is mistakenly classified as healthy [39].

In this study, the evaluation of the classification results is based on four key performance metrics derived from the confusion matrix: Accuracy, Precision, Recall, and the F1-Score. These measures provide a comprehensive assessment of the model's ability to correctly identify diabetic and non-diabetic cases.

Accuracy quantifies the ratio of correctly classified instances to the total number of instances, thereby indicating the overall efficacy of the classifier.

$$Accuracy = \frac{TP + TN}{TP + TN + FP + FN} \quad (1)$$

Precision quantifies the ratio of true positives to the total number of instances predicted as positive.

$$Precision = \frac{TP}{TP + FP} \quad (2)$$

Recall quantifies the ratio of true positives to the total number of actual positive instances.

$$Recall = \frac{TP}{TP + FN} \quad (3)$$

F1-Score is the harmonic mean of Precision and Recall, providing a balance between them.

$$F1\text{-}Score = \frac{2TP}{2TP + FP + FN} \quad (4)$$

The upcoming section is dedicated to evaluating the results derived from the proposed model.

## IV. RESULTS

The proposed model in this study demonstrates a significant advancement by integrating three key innovations: the use of SMOTE-ENN to address class imbalance, the application of the Bat Algorithm not only for hyperparameter tuning but also for optimal feature selection, and the reduction of redundant features. Remarkably, with only six selected features, the model achieved 89% accuracy, 88% precision, 90% recall, and an F1-score of 89% (Table 2).

TABLE II. Comparison of the basic random forest model, our method model, and a related study.

| Algorithms | Number of Features | Accuracy (%) | Precision (%) | Recall (%) | F1-score (%) |
|---|---|---|---|---|---|
| **Random Forest** | 8 | 74 | 66 | 52 | 58 |
| **Anam et al.** [36] | 8 | 78 | 76 | 83 | 79 |
| **Verma & Khatoon** [33] | 8 | 80 | 86 | 84 | 85 |
| **Our Model** | 6 | 89 | 88 | 90 | 89 |

Compared to the baseline Random Forest, which with eight features yielded only 74% accuracy, 66% precision, 52% recall, and an F1-score of 58%, the proposed approach shows dramatic improvements, especially in recall—rising from 52% to 90%. This indicates that the model can identify diabetic patients much more reliably, significantly reducing false negatives, a critical concern in medical diagnosis. Moreover, when benchmarked against prior studies such as Anam et al. (2025) [36] (accuracy 78%, F1-score 79%) and Verma & Khatoon (2024) [33] (accuracy 80%, F1-score 85%), our model clearly outperforms them in all major evaluation metrics while requiring fewer input features.

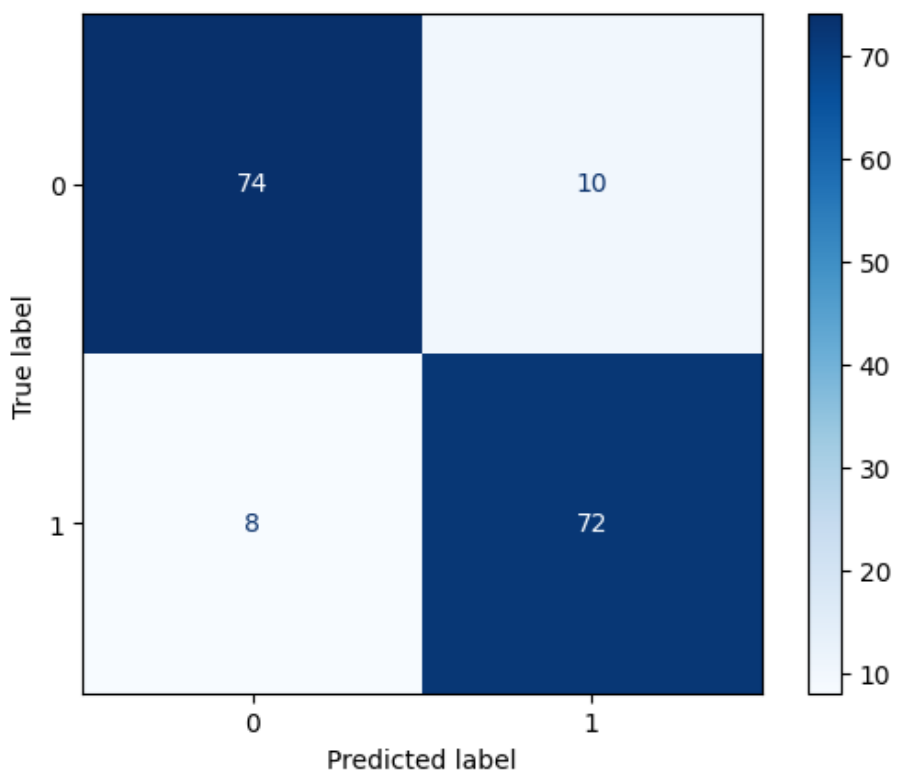

Fig. *2*. Our *model* confusion matrix.

The confusion matrix shown in Figure 2 further illustrates this improvement, with far fewer misclassifications and a clearer separation between diabetic and non-diabetic patients. Taken together, these results emphasize that while Random Forest alone is insufficient for reliable clinical prediction, optimization with swarm intelligence methods like the Bat Algorithm can greatly enhance performance. Beyond that, the combined strategy of data balancing, hyperparameter tuning, and feature selection proposed in this study yields a model that is both more efficient and significantly more accurate. The reduction in the number of required features without sacrificing—indeed, while improving—predictive power suggests that the model could serve as a practical, cost-effective, and highly accurate tool for early diabetes diagnosis, with strong potential for real-world medical applications.

## V. Conclusion

In this study, we proposed a model for type 2 diabetes diagnosis that integrates three interrelated components. First, systematic preprocessing was applied to the Pima Indians Diabetes Database, including the correction of physiologically invalid values (hidden zeros), detection and removal of outliers, and normalization to ensure a clean and stable dataset for learning. This benchmark dataset consists of 768 records and eight clinical features and served as the foundation of our analysis. Second, class imbalance was addressed using the SMOTE-ENN technique, which combines synthetic oversampling with the removal of ambiguous samples to reduce bias and improve model generalization. Third, the Bat Optimization Algorithm was employed for simultaneous feature selection and hyperparameter tuning of the Random Forest classifier, enabling the exclusion of redundant features and the adjustment of key parameters such as the number and depth of trees to prevent overfitting or underfitting. This threefold integration significantly enhanced performance: while the baseline Random Forest achieved only 74% accuracy, 52% recall, and an F1-score of 58%, the optimized model reached 89% accuracy, 88% precision, 90% recall, and an F1-score of 89% using only six selected features

Despite these promising results, important limitations must be acknowledged. The Pima dataset represents a single-center study, restricted to a specific demographic group—women of Pima Indian heritage in Arizona—thereby limiting its diversity in terms of genetics, lifestyle, and environmental factors. Moreover, its relatively small sample size and the limited set of clinical features constrain the model's ability to capture the broader complexity of type 2 diabetes. The dataset also contains outdated records and implausible zero entries, which require careful preprocessing and may affect the robustness of the findings.

Future work should therefore focus on validating the proposed model with larger, multi-center, and more up-to-date clinical datasets. Potential directions include enhancing interpretability through methods such as SHAP, exploring alternative metaheuristics or hybrid deep learning models, leveraging continuous glucose monitoring (CGM) and temporal data for richer predictions, and assessing the clinical cost of false negatives in real-world settings. Overall, the proposed model demonstrates the potential of combining evolutionary algorithms with ensemble learning to deliver a practical, cost-effective, and accurate tool for early type 2 diabetes screening, paving the way for decision-support systems in clinical practice.

## References

[1] R. Zhao, Z. Lu, J. Yang, L. Zhang, Y. Li, and X. Zhang, "Drug Delivery System in the Treatment of Diabetes Mellitus," *Front. Bioeng. Biotechnol.*, vol. 8, 2020, doi: 10.3389/fbioe.2020.00880.

[2] S. M. Ju, M. Y. Ali, S. M. Ko, J. H. Ryu, J. S. Choi, and H. A. Jung, "An Arylbenzofuran, Stilbene Dimers, and Prenylated Diels–Alder Adducts as Potent Diabetic Inhibitors from Morus bombycis Leaves," *Antioxidants*, vol. 12, no. 4, p. 837, 2023, doi: 10.3390/antiox12040837.

[3] D. Mohajan and H. K. Mohajan, "Historical View of Diabetics Mellitus: From Ancient Egyptian Polyuria to Discovery of Insulin," *Stud. Soc. Sci. Humanit.*, vol. 2, no. 7, pp. 26–34, 2023, doi: 10.56397/sssh.2023.07.05.

[4] S. P. Manalu and T. N. Nurzanna, "Conducting Counseling About Diabetes Mellitus in Desa Perkebunan Tanjung Kasau. Kec, Laut Tador. Kab, Batu Bara," *ABDIMAS Talent. J. Pengabdi. Kpd. Masy.*, vol. 8, no. 2, pp. 893–899, 2023, doi: 10.32734/abdimastalenta.v8i2.10619.

[5] J. Huang *et al.*, "Artificial Intelligence for Predicting and Diagnosing Complications of Diabetes," *J. Diabetes Sci. Technol.*, vol. 17, no. 1, pp. 224–238, 2023, doi: 10.1177/19322968221124583.

[6] S. M. Dsouza *et al.*, "Effectiveness of self-management applications in improving clinical health outcomes and adherence among diabetic individuals in low and middle-

income countries: a systematic review," *BMJ Open*, vol. 12, no. 11, p. e060108, 2022, doi: 10.1136/bmjopen-2021-060108.

[7] I. El Ouarti and F. Abdallaoui, "Unusual Periradicular Lesion Shape with Endodontic Origin in a Diabetic Patient: A Case Report," *Integr. J. Med. Sci.*, 2022, doi: 10.15342/ijms.2022.571.

[8] A. Dhore, "Gestational Diabetes Mellitus Among Pregnant Women: A Review," *Indian J. Clin. Cardiol.*, vol. 5, no. 4, pp. 403–403, 2024, doi: 10.1177/26324636241283178.

[9] K. M. Aamir, L. Sarfraz, M. Ramzan, M. Bilal, J. Shafi, and M. Attique, "A fuzzy rule-based system for classification of diabetes," *Sensors*, vol. 21, no. 23, p. 8095, 2021, doi: 10.3390/s21238095.

[10] S. F. Sajjadi, J. W. Sacre, B. Carstensen, S. Ruiz-Carmona, J. E. Shaw, and D. J. Magliano, "Evaluating the incidence of complications among people with diabetes according to age of onset: Findings from the UK Biobank," *Diabet. Med.*, vol. 41, no. 9, 2024, doi: 10.1111/dme.15349.

[11] I. S. Jeong and C. M. Kang, "Time to Diagnosis and Treatment of Diabetes Mellitus among Korean Adults with Hyperglycemia: Using a Community-Based Cohort Study," *Int. J. Environ. Res. Public Health*, vol. 19, no. 19, p. 12090, 2022, doi: 10.3390/ijerph191912090.

[12] M. Auzanneau, A. Fritsche, A. J. Eckert, E. Seidel-Jacobs, M. Heni, and S. Lanzinger, "Advantages of DDG-certified Hospitals for Hospitalized Patients With Diabetes - A Nationwide DRG Analysis in Germany," 2025, doi: 10.1101/2025.04.07.25325368.

[13] Z. Wang, S. Wang, M. Wang, and Y. Sun, "Design of application-oriented disease diagnosis model using a meta-heuristic algorithm," *Technol. Heal. Care*, vol. 32, no. 6, pp. 4041–4061, 2024, doi: 10.3233/THC-231755.

[14] A. Champakanath, H. K. Akturk, G. T. Alonso, J. K. Snell-Bergeon, and V. N. Shah, "Continuous Glucose Monitoring Initiation Within First Year of Type 1 Diabetes Diagnosis Is Associated With Improved Glycemic Outcomes: 7-Year Follow-Up Study," *Diabetes Care*, vol. 45, no. 3, pp. 750–753, 2022, doi: 10.2337/dc21-2004.

[15] J. Golding, S. V. Hope, A. J. Chakera, and A. Puttanna, "The evolving continuum of dysglycaemia: Non-diabetic hyperglycaemia in older adults," *Diabet. Med.*, vol. 40, no. 10, 2023, doi: 10.1111/dme.15177.

[16] B. Zhou *et al.*, "Global variation in diabetes diagnosis and prevalence based on fasting glucose and hemoglobin A1c," *Nat. Med.*, vol. 29, no. 11, pp. 2885–2901, 2023, doi: 10.1038/s41591-023-02610-2.

[17] M. Ohigashi *et al.*, "Association of time in range with hemoglobin A1c, glycated albumin and 1,5-anhydro-d-glucitol," *J. Diabetes Investig.*, vol. 12, no. 6, pp. 940–949, 2021, doi: 10.1111/jdi.13437.

[18] Y. Shen *et al.*, "Association between visit-to-visit HbA1c variability and the risk of cardiovascular disease in patients with type 2 diabetes," *Diabetes, Obes. Metab.*, vol. 23, no. 1, pp. 125–135, 2021, doi: 10.1111/dom.14201.

[19] Y. Wen and Q. Wang, "Cut-off value of glycated hemoglobin A1c for detecting diabetic retinopathy in the Chinese population," *World J. Diabetes*, vol. 15, no. 7, pp. 1531–1536, 2024, doi: 10.4239/wjd.v15.i7.1531.

[20] W. Yu *et al.*, "Opportunistic screening for diabetes among urban Aboriginal adults during emergency department attendance," *Intern. Med. J.*, vol. 55, no. 7, pp. 1146–1151, 2025, doi: 10.1111/imj.70078.

[21] Y. Yan *et al.*, "Predictive Ability of Visit-to-Visit Variability of HbA1c Measurements for the Development of Diabetic Kidney Disease: A Retrospective Longitudinal Observational Study," *J. Diabetes Res.*, vol. 2022, pp. 1–11, 2022, doi: 10.1155/2022/6934188.

[22] C. Wei, "Application of Machine Learning in the Field of Diabetes," *Sci. Technol. Eng. Chem. Environ. Prot.*, vol. 1, no. 8, 2024, doi: 10.61173/kc05df27.

[23] S. Ziajor *et al.*, "The use of artificial intelligence in the diagnosis and detection of complications of diabetes," *J. Educ. Heal. Sport*, vol. 65, pp. 11–27, 2024, doi: 10.12775/jehs.2024.65.001.

[24] D. D. Rufo, T. G. Debelee, A. Ibenthal, and W. G. Negera, "Diagnosis of diabetes mellitus using gradient boosting machine (Lightgbm)," *Diagnostics*, vol. 11, no. 9, p. 1714, 2021, doi: 10.3390/diagnostics11091714.

[25] J. S. Wekesa and M. Kimwele, "A review of multi-omics data integration through deep learning approaches for disease diagnosis, prognosis, and treatment," *Front. Genet.*, vol. 14, 2023, doi: 10.3389/fgene.2023.1199087.

[26] V. Chang, J. Bailey, Q. A. Xu, and Z. Sun, "Pima Indians diabetes mellitus classification based on machine learning (ML) algorithms," *Neural Comput. Appl.*, vol. 35, no. 22, pp. 16157–16173, Aug. 2023, doi: 10.1007/s00521-022-07049-z.

[27] J. Cai, J. Luo, S. Wang, and S. Yang, "Feature selection in machine learning: A new perspective," *Neurocomputing*, vol. 300, pp. 70–79, Jul. 2018, doi: 10.1016/j.neucom.2017.11.077.

[28] X. S. Yang and A. H. Gandomi, "Bat algorithm: A novel approach for global engineering optimization," *Eng. Comput. (Swansea, Wales)*, vol. 29, no. 5, pp. 464–483, Jul. 2012, doi: 10.1108/02644401211235834.

[29] A. F. H. Alharan, Z. M. Algelal, N. S. Ali, and N. Al-Garaawi, "Improving Classification Performance for Diabetes with Linear Discriminant Analysis and Genetic Algorithm," in *2021 Palestinian International Conference on Information and Communication Technology (PICICT)*, IEEE, Sep. 2021, pp. 38–44. doi: 10.1109/PICICT53635.2021.00019.

[30] R. Katarya and S. Jain, "Comparison of different machine learning models for diabetes detection," in *Proceedings of 2020 IEEE International Conference on Advances and Developments in Electrical and Electronics Engineering, ICADEE 2020*, IEEE, Dec. 2020, pp. 1–5. doi: 10.1109/ICADEE51157.2020.9368899.

[31] N. Abdulhadi and A. Al-Mousa, "Diabetes Detection Using Machine Learning Classification Methods," in *2021 International Conference on Information Technology, ICIT 2021 - Proceedings*, IEEE, Jul. 2021, pp. 350–354. doi: 10.1109/ICIT52682.2021.9491788.

[32] I. H. Sarker, "Machine Learning: Algorithms, Real-World Applications and Research Directions," *SN Comput. Sci.*, vol. 2, no. 3, p. 160, May 2021, doi: 10.1007/s42979-021-00592-x.

[33] P. Verma and A. Khatoon, "Data Mining Applications in Healthcare: A Comparative Analysis of Classification Techniques for Diabetes Diagnosis Using the PIMA Indian Diabetes Dataset," in *2024 4th International Conference on Innovative Practices in Technology and Management (ICIPTM)*, IEEE, Feb. 2024, pp. 1–5. doi: 10.1109/ICIPTM59628.2024.10563296.

[34] S. Sreejith, H. Khanna Nehemiah, and A. Kannan, "Clinical data classification using an enhanced SMOTE and chaotic evolutionary feature selection," *Comput. Biol. Med.*, vol. 126, p. 103991, Nov. 2020, doi: 10.1016/j.compbiomed.2020.103991.

[35] Z. Mushtaq, M. F. Ramzan, S. Ali, S. Baseer, A. Samad, and M. Husnain, "Voting Classification-Based Diabetes Mellitus Prediction Using Hypertuned Machine-Learning Techniques," *Mob. Inf. Syst.*, vol. 2022, pp. 1–16, Mar. 2022, doi: 10.1155/2022/6521532.

[36] S. Anam, F. Deny Tisna Amijaya, S. Hadi Wijoyo, D. Eka Ratnawati, C. Ayu Dwi Lestari, and M. Ilyas, "Diabetes mellitus diagnosis method based random forest with bat algorithm," *IAES Int. J. Artif. Intell.*, vol. 14, no. 2, p. 1140, Apr. 2025, doi: 10.11591/ijai.v14.i2.pp1140-1149.

[37] R. Saxena, S. K. Sharma, M. Gupta, and G. C. Sampada, "A Comprehensive Review of Various Diabetic Prediction Models: A Literature Survey," *J. Healthc. Eng.*, vol. 2022, pp. 1–15, Apr. 2022, doi: 10.1155/2022/8100697.

[38] G. Husain *et al.*, "SMOTE vs. SMOTEENN: A Study on the Performance of Resampling Algorithms for Addressing Class Imbalance in Regression Models," *Algorithms*, vol. 18, no. 1, p. 37, Jan. 2025, doi: 10.3390/a18010037.

[39] A. U. Haq *et al.*, "Intelligent Machine Learning Approach for Effective Recognition of Diabetes in E-Healthcare Using Clinical Data," *Sensors*, vol. 20, no. 9, p. 2649, May 2020, doi: 10.3390/s20092649.